\documentclass[10pt, a4paper]{article}
\usepackage{lrec2022} 
\usepackage{multibib}
\usepackage{graphicx}
\usepackage{tabularx}
\usepackage{soul}
\usepackage{titlesec}
\titleformat{\section}{\normalfont\large\bfseries\center}{\thesection.}{1em}{}
\titleformat{\subsection}{\normalfont\SmallTitleFont\bfseries\raggedright}{\thesubsection.}{1em}{}
\titleformat{\subsubsection}{\normalfont\normalsize\bfseries\raggedright}{\thesubsubsection.}{1em}{}
\renewcommand\thesection{\arabic{section}}
\renewcommand\thesubsection{\thesection.\arabic{subsection}}
\renewcommand\thesubsubsection{\thesubsection.\arabic{subsubsection}}

\usepackage{epstopdf}
\usepackage[utf8]{inputenc}

\usepackage{hyperref}
\usepackage{xstring}

\usepackage{color}

\usepackage{times}
\usepackage{latexsym}
\usepackage{danudefs}
\usepackage{algorithmic}
\usepackage{algorithm}
\usepackage{booktabs}
\usepackage{xspace}
\usepackage{url}
\usepackage{mathtools}
\usepackage{amsmath}
\usepackage{graphicx}
\usepackage{subfigure} 
\usepackage{multirow}
\usepackage{enumerate}
\usepackage{esvect}
\usepackage{csvsimple}
\usepackage{microtype}
\usepackage{enumitem}

\usepackage{todonotes}
\usepackage{threeparttable}

\makeatletter
\if@todonotes@disabled

\else

\fi
\makeatother

\usepackage[english]{babel}
\usepackage{hyperref}
\addto\captionsenglish{%
}
\addto\extrasenglish{%
}

\title{Unsupervised Attention-based Sentence-Level Meta-Embeddings from Contextualised Language Models}

\name{$\textrm{Keigo Takahashi}^{1}$, $\textrm{Danushka Bollegala}^{2,3}$\thanks{Danushka Bollegala holds concurrent appointments as a Professor at University of Liverpool and as an Amazon Scholar. This paper describes work performed at the University of Liverpool and is not associated with Amazon.}}

\address{$\textrm{Tokyo Metropolitan University}^{1}$, $\textrm{University of Liverpool}^{2}$, $\textrm{Amazon}^{3}$ \\
         takahashi-keigo@ed.tmu.ac.jp, danushka@liverpool.ac.uk}

\abstract{
A variety of contextualised language models have been proposed in the NLP community, which are trained on diverse corpora to produce numerous Neural Language Models (NLMs).
However, different NLMs have reported different levels of performances in downstream NLP applications when used as text representations.
We propose a sentence-level meta-embedding learning method that takes independently trained contextualised word embedding models and learns a sentence embedding that preserves the complementary strengths of the input source NLMs.
Our proposed method is unsupervised and is not tied to a particular downstream task, which makes the learnt meta-embeddings in principle applicable to different tasks that require sentence representations.
Specifically, we first project the token-level embeddings obtained by the individual NLMs and learn attention weights that indicate the contributions of source embeddings towards their token-level meta-embeddings.
Next, we apply mean and max pooling to produce sentence-level meta-embeddings from token-level meta-embeddings.
Experimental results on semantic textual similarity benchmarks show that our proposed unsupervised sentence-level meta-embedding method outperforms previously proposed sentence-level meta-embedding methods as well as a supervised baseline.
 \\ \newline  }

\begin{document}

\maketitleabstract

\section{Introduction}
\label{sec:intro}





Meta-embedding~\cite{Yin:ACL:2016,Bollegala:IJCAI:2018} is the task of producing embeddings from a given set of pretrained text embeddings such that the complementary strengths of the input (source) embeddings are preserved or further enhanced to obtain the best performance on a downstream NLP task.
Numerous methods have been proposed for creating a meta-embedding from a given set of static word embeddings such as GloVe \cite{Pennington:EMNLP:2014}, Skip-gram with Negative Sampling (SGNS) \cite{NIPS2013_9aa42b31} or Continuous Bag-of-Words (CBOW)  \cite{DBLP:journals/corr/abs-1301-3781} as the source embeddings.
In contrast to static word embeddings, contextualised word embeddings obtained from Neural Language Models (NLMs) such as BERT \cite{BERT}, RoBERTa \cite{RoBERTa} etc. represent the same word with different vectors in different contexts, thereby providing a richer context-dependent semantic representations.
Moreover, contextualised word embeddings can be pooled and fine-tuned to obtain accurate sentence-level representations.
Compared to static word embeddings, contextualised word/sentence embeddings have reported impressive performances in a wide range of NLP tasks.

\paragraph{Challenges:}
Compared to meta-embedding of static word embeddings, when computing meta-embeddings from contextualised word embeddings~\cite{poerner-etal-2020-sentence} one must overcome several challenges.
\begin{itemize}
 \item First, unlike static word embeddings, which represent the same word using the same embedding vector in all contexts in which the target word occurs, contextualised word embeddings depend on the context of the word.
Therefore, the same word will be represented by different embeddings in different contexts by a contextualised word embedding model.
This is particularly challenging when creating meta word embeddings because the meta embedding of a word will depend not only on the source embeddings, but also on the context in which the target word occurs.

\item Second, static word embeddings represent words using relatively smaller dimensional vector spaces compared to contextualised word embeddings.
For example, good performance can be obtained with as small as 300-600 dimensional GloVe static word embeddings, whereas BERT contextualised word embeddings require 768, 1024 or higher dimensionalities.
Therefore, methods such as concatenation that have been effective for creating meta-embeddings~\cite{coates-bollegala-2018-frustratingly} for static word embeddings become problematic when applied to contextualised word embeddings.
This issue is further aggravated by the fact that contextualised word embedding models require specialised hardware such as GPUs, which have limited memory buffers.

\item Third, although supervised approaches have been proposed for meta-embedding of static word embeddings that require labelled data for the target tasks~\cite{xie2019dynamic,he-etal-2020-learning,wu2020task}, adapting such supervised approaches for contextualised word embeddings remains a challenging task.
Because contextualised word embeddings such as BERT typically contain over 100 million parameters,\footnote{For example, the \texttt{BERT-small} model trained by \newcite{BERT} with 12 layers contain 110 million parameters}
it requires significantly more labelled data to fine-tune such models for a given target task.
Therefore, unsupervised methods are preferable over supervised ones when creating meta-embeddings from contextualised embedding models.
\end{itemize}

\paragraph{Proposal:}
To address the above-mentioned challenges, we propose an unsupervised method\footnote{Codes: \url{https://github.com/keigotak/unsup-att-based-meta-sent-emb}} to combine multiple contextualised word embedding models (referred to as \textbf{source embeddings}) to create a meta-embedding for a given target word in a sentence that it occurs.
Furthermore, we create sentence-level embeddings by applying parameter-free operations such as \emph{pooling} to create sentence-level meta-embeddings from the source embeddings.

In our proposed method, we first project token-level embeddings obtained for a word from a contextualised word embedding to a common meta-embedding space using linear transformations. 
We then compute the linearly-weighted average of the projected source embeddings to compute their token-level meta-embedding.
The weighting coefficients can be seen as describing the significance we must impart on a source embedding in the ensemble.
Note that contextualised token-embeddings produced by different NLMs can in general have different dimensionalities. 
Therefore, the source-specific linear projections learnt in the first step ensures all source embeddings are projected to a common meta-embedding dimensionality such that we can compute their linearly-weighted average.
Finally, we apply pooling operations such as max and mean pooling over the token-level meta-embeddings for the words in a sentence to create a sentence-level meta-embedding for the corresponding sentence.

\paragraph{Contributions:}
We evaluate our proposed meta embedding learning method using different types of source embeddings on Semantic Textual Similarity (STS) benchmarks \cite{cer-etal-2017-semeval}.
We compare our results against multiple baselines such as vector concatenation, averaging, singular value decomposition and the current state-of-the-art unsupervised sentence-level meta-embedding method proposed by \newcite{poerner-etal-2020-sentence}.
Experimental results show that our proposed unsupervised sentence-level meta-embedding learning method obtains the highest correlation with human similarity ratings in all STS benchmark datasets, outperforming all methods compared.

\section{Related Work}
\subsection{Sentence-Level Meta-embeddings}
Sentence-level meta-embedding learning methods aim to create a single vector representation for a given sentence by combining token-level representations produced by multiple source embeddings.
\newcite{kiela-etal-2018-dynamic} proposed dynamic meta-embeddings, which learns task-specific combinations of source embeddings using the attention mechanism~\cite{Bahdanau:2014}.
\newcite{poerner-etal-2020-sentence} proposed the use of Generalised Canonical Correlation Analysis (GCCA) to learn sentence-level meta-embeddings from contextualised word embeddings.
GCCA is an extension of Canonocal Correlation Analysis (CCA) to more than two random variables.
Specifically, a word in a sentence can be represented using multiple vectors by using multiple contextualised word embedding models.
GCCA learns linear transformations that can be applied to such vectors to maximise the correlation in the projected space.

Unlike our proposed method, prior work on sentence-level meta-embedding methods consider sentence embeddings as the input to the meta-embedding algorithms.
In contrast, we first create token-level meta-embeddings by combining multiple contextualised embeddings and then create sentence-level meta-embeddings by applying pooling operations (e.g. max and min pooling) over the words in a sentence.
GCCA-based meta sentence embedding method proposed by \newcite{poerner-etal-2020-sentence} is the current state-of-the-art unsupervised sentence-level meta-embedding learning method and is a direct comparison to ours which is also unsupervised method.
As we later see in \autoref{sec:exp}, our proposed method outperforms GCCA in multiple benchmark datasets.

\subsection{Word-Level Meta-embeddings}
Word meta-embedding has been studied more actively compared to sentence-level meta-embedding.
\newcite{Yin:ACL:2016} proposed 1\texttt{TO}N,
 by projecting source embeddings to a common space via source-specific linear transformations.
This method minimises squared $\ell_{2}$ distance between the meta and each source embedding assuming a common vocabulary. 
1\texttt{TO}N+ overcomes this limitation by learning pairwise linear transformations between two given sources for predicting the embeddings for out of vocabulary (OOV) words.
Both of these methods can be seen as \emph{globally-linear} transformations because \emph{all} the words in a particular source are projected to the meta-embedding space using the \emph{same} transformation matrix.

In contrast, \emph{locally-linear} meta-embedding (LLE)~\cite{Bollegala:IJCAI:2018} computes the nearest neighbours for a particular word in each source and then represent a word as the linearly-weighted combination of its neighbours.
Next, meta-embeddings are learnt by preserving the neighbourhood constraints.
This method does not require all words to be represented by all sources, thereby obviating the need to predict missing source embeddings for OOVs.

\newcite{Bao:COLING:2018} modelled meta-embedding learning as an \emph{autoencoding} problem where information embedded in different sources are integrated at different levels to propose three types of meta-embeddings: Decoupled Autoencoded Meta-Embedding (DAEME) (independently encode each source and  concatenate), Concatenated Autoencoded Meta-Embedding (CAEME) (independently decode meta-embeddings to reconstruct each source), and Averaged Autoencoded Meta-Embedding (AAEME) (similar to DAEME but instead of concatenation use average).
In comparison to methods that learn globally or locally linear transformations~\cite{Bollegala:IJCAI:2018,Yin:ACL:2016}, autoencoders learn nonlinear transformations.
\newcite{Neill:2018} further extend this approach using squared cosine proximity loss as the reconstruction loss. 

Vector concatenation has been used as a baseline for producing 
meta-embeddings~\cite{Bao:COLING:2018,Yin:ACL:2016,Bollegala:IJCAI:2018}.
\newcite{AAAI:2016:Goikoetxea} concatenated independently learnt word embeddings from a corpus and the WordNet.
Moreover, applying Principal Component Analysis on the concatenation further improved their performance on similarity tasks.
Interestingly, \newcite{Coates:NAACL:2018} showed that accurate meta-embeddings can be produced by averaging source embeddings that exist in \emph{different} vector spaces. 
Specifically, they showed that when word embeddings in each source are approximately orthogonal (they empirically validate for pre-trained embeddings), averaging can approximate concatenation. 
Although averaging does not increase the dimensionality of the meta-embedding space as with concatenation, it does not consistently outperform concatenation, especially when the orthogonality condition does not hold.
Recent work~\cite{He:2020,jawanpuria-etal-2020-learning} show that learning orthogonal transformations prior to averaging can further improve accuracy.

\section{Method}
\label{sec:unsup}

Let us denote a sentence $s = w_1, w_2, \ldots, w_{|s|}$ consisting of tokens $w_1, w_2, \ldots, w_{|s|}$ in that order.
In the case where we have applied a subtokenisation method such as Byte Pair Encoding \newcite{Sennrich:2016} to tokenise sentences into subtokens, we will assume $w_j$ to be subtokens instead of tokens. 
In the following discussion, we use the term \emph{word} to cover both tokens as well as subtokens unless otherwise explicitly stated.
Moreover, let us assume that we are given $n$ independently pretrained contextualised word embeddings of which the $i$-th source embedding is denoted by $\vec{f}_i (w, s) \in \R^{d_{i}}$, and returns a $d_i$ dimensional word embedding of $w$ in a given context $s$ (e.g. a sentence or a contextual window).
In particular, static word embeddings can be seen as a special case of this formulation where the word embedding does not depend on $s$ (i.e. $\forall s$ $\vec{f}_i(w,s) = \vec{f}_i(w)$).
Our goal in this paper is to create a sentence-level meta-embedding, $\vec{m}(s)$, representing the sentence $s$ using all word-level source embeddings, $\{\vec{f}_i(w,s)\}_{i=1}^{n}$, for the words $w$ in $s$. 
Following the convention used in prior work on meta-embedding learning, we refer to individual $\vec{f}_i$ as source embeddings.

Note that in general the individual source embeddings will have different dimensionalities $d_i$.
To transform each source embedding to the same meta-embedding space with dimensionality $d_m$, we first transform each source embedding by a source-specific projection matrix $\mat{A}_i \in \R^{d_m \times d_i}$ as given by \eqref{eq:wo-attention}.
\begin{align}
    \label{eq:wo-attention}
    \vec{\phi}_i(w,s) = \mat{A}_i \vec{f}_i(w,s)
\end{align}
Here, the projected source embeddings, $\vec{\phi}(w,s) \in \R^{d_m}$, will be in the same common $d_m$ dimensional meta-embedding space for all source embeddings.
The total number of learnable parameters in this model is $\sum_{i=1}^n d_m d_i$, which corresponds to the sum of elements in all projection matrices $\{\mat{A}_i\}_{i=1}^{n}$.

Different source embeddings will have complementary strengths and weaknesses. 
Therefore, we propose to assign weights to the source embeddings when creating a meta-embedding of those source embeddings as given in \eqref{eq:w-attention}.
\begin{align}
    \label{eq:w-attention}
    \vec{\phi}_i(w_j,s_k) = \alpha_{i} \mat{A}_i \vec{f}_i(w_j,s_k)
\end{align}
Here, $\alpha_i$ is the weight associated with the $i$-th source embedding such that $\sum_i \alpha_i = 1$.
Moreover, to reduce the degrees of freedom of the model, we require cross-correlations in projection matrices to be small.
This requirement can be implemented as a regularisation on the projection matrices such that $\norm{\mat{A}_i\T\mat{A}_i - \mat{I}}_F^2$ is minimised, where $\norm{\cdot}_F$ denotes the Frobenius norm.

We can consider the individual projection matrices $\mat{A}_i$ to be concerned with projecting source embeddings with different dimensionalities to a common meta-embedding space, whereas $\alpha_i$ determines the weight imparted on the projected source embedding in the meta-embedding space.
However, in the unsupervised meta-embedding learning setting that we consider in this paper, we do not assume the availability of any labelled data for downstream tasks nor our goal is to learn a meta-embedding that is specific to a particular task.
On the other hand, task specialisation of meta-embedding can be done as a subsequent task akin to fine-tuning of foundation models.
Therefore, we learn both $\alpha_i$ and $\mat{A}_i$ via an unsupervised learning method as described later.

Given $n$ distinct source embeddings, $\{\vec{f}_i(w,s)\}_{i=1}^{n}$, of a word $w$ in a sentence $s$, we compute the word's meta-embedding, $\vec{m}_{\rm word}(w,s)$, as the average of the projected source embeddings as given by \eqref{eq:meta-word-embedding}.
\begin{align}
    \label{eq:meta-word-embedding}
    \vec{m}_{\rm word}(w,s) = \frac{1}{n} \sum_{i=1}^{n} \vec{\phi}_i(w,s)
\end{align}
Recall that the projection matrices ensure that source embeddings with different dimensionalities are projected to the same $d_m$ dimensional meta-embedding space such that they can be averaged to obtain a reliable estimate of the final word-level meta-embedding.

Armed with the word-level meta-embeddings, $\vec{m}_{\rm word}(w,s)$, for a word $w$ in a sentence $s$, we compute two types of  sentence-level meta-embeddings, $\vec{m}_{\rm mean}(s)$ (defined by \eqref{eq:max-pool}) and $ \vec{m}_{\rm max}(s)$ (defined by \eqref{eq:mean-pool}) by applying respectively mean or max pooling  over the word embeddings $\vec{m}_{\rm word}(w,s)$ corresponding to the words in the sentence $s$.
\begin{align}
     &\vec{m}_{\rm mean}(s) = \frac{1}{|s|} \sum_{w \in s} \vec{m}_{\rm word}(w,s) \label{eq:mean-pool} \\
    &\vec{m}_{\rm max}(s) = \max_{w \in s} \vec{m}_{\rm word}(w,s) \label{eq:max-pool} 
\end{align}
The pooling operations are conducted elementwise over the set of word-level meta-embeddings.
We compare the performances of meta sentence embeddings created using mean and max pooling in \autoref{sec:exp}.

What remains in our model is to devise a method to learn the projection matrices, $\{\mat{A}_i\}_{i=1}^{n}$, and the weights $\{\alpha_i\}_{i=1}^{n}$.
Recall that in this paper we are considering \emph{unsupervised} sentence embeddings.
Therefore, we do not assume the availability of labelled data such as human-annotated similarity ratings used in supervised sentence embedding learning.
For this reason, next we derive an unsupervised training objective.

Let us consider the projected source embeddings of two words $w \in s$, $w' \in s'$ in two different sentences $s$ and $s'$ to be given respectively by $\vec{\phi}_i(w,s)$ and $\vec{\phi}_j(w',s')$ using two different sources $f_i$ and $f_j$.
We assume the following four criteria to be satisfied by these projected source embeddings.
\begin{description}
\item [Criterion 1:] \emph{The projected source embeddings $\vec{\phi}_i(w,s)$ and $\vec{\phi}_j(w,s)$ of the same word in the same sentence must be similar.}

This requirement stems from the definition of meta-embedding process because the purpose of the projection matrix in the first place was to ensure that different source embeddings of the same word are closer to each other in the meta-embedding space. 
This requirement can be formulated as the minimisation of the squared $\ell_2$ distance between the projected source embeddings given by \eqref{eq:crit-1}.
\begin{align}
    \label{eq:crit-1}
    \norm{\vec{\phi}_i(w,s) - \vec{\phi}_j(w,s)}_2^2
\end{align}

\item [Criterion 2:] \emph{The projected source embeddings $\vec{\phi}_i(w,s)$ and $\vec{\phi}_j(w',s)$ of two different words $w$ and $w'$ that co-occur in the same sentence $s$ must be dissimilar.}

This requirement is important to make the meta-embeddings of two distinct words to be sufficiently dissimilar such that we can discriminate them in the meta-embedding space.
We formalise this requirement as the minimisation of the squared $\ell_2$ distance given by \eqref{eq:crit-2}
\begin{align}
    \label{eq:crit-2}
    \norm{\vec{\phi}_i(w,s) - \vec{\phi}_j(w',s)}_2^2
\end{align}

\item [Criterion 3:] The projected source embeddings $\vec{\phi}_i(w,s)$ and $\vec{\phi}_j(w,s')$ of the same word $w$ in two different sentences $s$ and $s'$ must be dissimilar.

This is particularly true for contextualised source embeddings, which consider both the word as well as the context in which the word occurs when representing the word.
Moreover, we are considering embeddings produced by two distinct source embeddings here.
This requirement is useful for creating different meta-embeddings for different sentences eventhough they might share some words in common. 
We formalise this requirement as the minimisation of the squared $\ell_2$ distance given by \eqref{eq:crit-3}.
\begin{align}
    \label{eq:crit-3}
    \norm{\vec{\phi}_i(w,s) - \vec{\phi}_j(w,s')}_2^2
\end{align}

\item [Criterion 4:] \emph{The projected source embeddings $\vec{\phi}_i(w,s), \vec{\phi}_j(w',s')$ of two different words $w, w'$ each occurring in different sentences $s, s'$ must be dissimilar.}

This requirement can be seen as the most extreme dissimilar case involving distinct sources, distinct sentences and distinct words.
It can also be seen as imposing a regularisation on the parameter learning objective that prevents degenerated solutions such as projecting all words and sentences to the same point in the meta-embedding space.
We formalise this requirement as the minimisation of the squared $\ell_2$ loss given by \eqref{eq:crit-4}.

\begin{align}
    \label{eq:crit-4}
    \norm{\vec{\phi}_i(w,s) - \vec{\phi}_j(w',s')}_2^2
\end{align}
\end{description}

Finally, we combine the above-mentioned four constraints into a single training objective including the regulariser for the projection matrices as in \eqref{eq:unsup-loss}.
\begin{align}
    \label{eq:unsup-loss}
    &L(\{\mat{A}_i\}_{i=1}^{n}, \{\alpha_i\}_{i=1}^{n}) \nonumber \\ 
    =& \norm{\vec{\phi}_i(w,s) - \vec{\phi}_j(w,s)}_2^2 \nonumber \\
    & - \lambda \norm{\vec{\phi}_i(w,s) - \vec{\phi}_j(w',s)}_2^2 \nonumber \\
    & - \mu \norm{\vec{\phi}_i(w,s) - \vec{\phi}_j(w,s')}_2^2 \nonumber \\
    & - \nu \norm{\vec{\phi}_i(w,s) - \vec{\phi}_j(w',s')}_2^2 \nonumber \\
    &+ \xi \sum_{i=1}^{n}\norm{\mat{A}_i\T\mat{A}_i - \mat{I}}_F^2
\end{align}
Here, the regularisation coefficients $\lambda$, $\mu$, $\nu$ and $\xi$ are all optimised during training via backpropagation.

We randomly initialise all projection matrices and attention vectors and use Stochastic Gradient Descent (SGD) to optimise.
Further details on learning settings are provided in \autoref{sec:hyperparameters}.

Note that we do not assume any constraints for the projections of a single source embedding such as for two words in the same sentence, or the same word in different sentence for two reasons.
First, many source embedding learning methods already use these constraints during pretraining. 
For example, masked language models such as BERT, masks a word in a sentence and requires that it can be predicted correctly using the embeddings of the other words in the sentence. 
This requirement forces the words in the same sentence to have similar embeddings.
Second, during meta-embedding learning, we would like to preserve as much information as possible in the original sources, without imposing constraints for the individual sources, beyond what they have been already trained with.

Note that we require no human labelled data such as labels for sentences or similarity ratings for creating sentence-level meta-embedding from the above-described proposed method.
Therefore, we denote our unsupervised proposed method by \textbf{UNSUP} in the remainder of the paper.

\section{Experiments}
\label{sec:exp}

\subsection{Source Embeddings}

We use two source embeddings in our experiments -- 
BERT\footnote{1024-dimensional token embeddings. Labeled as \texttt{bert-large-nil-stsb-mean-tokens} in sBERT.} and
RoBERTa.\footnote{1024-dimensional token embeddings. Labeled as \texttt{roberta-large-nil-stsb-mean-tokens} in sBERT.}
All the models are pretrained by \newcite{reimers-2019-sentence-bert} and \newcite{wolf-etal-2020-transformers}.
We used the final layer of source embeddings for creating sentence representations. 
We note that the proposed sentence-level meta-embedding method can be used even without fine-tuning the source embeddings on sentence similarity related tasks.
However, the overall performance of meta-embeddings invariably depends on the quality of source embeddings used as the input to the meta-embedding method.
Therefore, it has been common practice in prior work on meta-embedding learning~\cite{poerner-etal-2020-sentence} to use high quality source embeddings as input.

Although we use two source embeddings as a proof-of-concept for the proposed method, note that the proposed method itself is not limited to only two sources. 
For example, given $n$ sources, we can consider pairwise combinations between any two of those $n$ sources to compute the four criteria described in \autoref{sec:unsup}.
Moreover, the BERT and RoBERTa sources we use in our evaluations have equal (i.e. 1024) dimensionalities.
However, the proposed method \emph{does not} require the dimensionalities of the source embeddings to be equal.
In fact, the projection matrices will ensure that the averaging operation given by \eqref{eq:meta-word-embedding} can be carried out even if the dimensionalities of the sources are different.
Further analysis on more than two sources and involving unequal dimensionalities are deferred to future work.


\subsection{Data and Evaluation Metrics}
\label{sec:data}
Given that our goal in this paper is to create sentence-level meta-embeddings, we use Semantic Textual Similarity (STS) prediction as an evaluation task.
In STS, we are given two sentences $s$ and $s'$ between which we must predict their semantic similarity.
For example, we can compute the cosine similarity between $s$ and $s'$ using the sentence embeddings computed using a meta sentence embedding method.
There exist several benchmark datasets where human annotators have provided similarity ratings, $r(s,s')$, between $s$ and $s'$, which can be considered as ground truth for evaluating sentence embedding methods.
We can then compute the agreements between human similarity ratings and the predicted cosine similarities using correlation measures such as the Spearman and Pearson correlations.

The SentEval tool~\cite{Conneau:2018ac} provides a unified framework for evaluating sentence embedding methods and reports the class-weighted averages of Pearson and Spearman correlation coefficients as the evaluation metric, which we use in this work.
A high degree of correlation between human similarity ratings and cosine similarity scores computed using the sentence embeddings produced by a meta sentence embedding method is considered as desirable, and accurately capturing the meaning of the sentences.
We use the following STS datasets in our experiments: STS-15~\cite{agirre-etal-2015-semeval}, STS-16~\cite{agirre-etal-2016-semeval}) and STS-B (STS benchmark)~\cite{cer-etal-2017-semeval}.

Note that the purpose of evaluating on STS benchmark dataset is \emph{not} to compare against the state-of-the-art for supervised sentence embedding methods but to compare against the unsupervised sentence-level meta-embedding learning methods.
With this objective in mind, we emphasise that the results reported in this paper must not be compared against general purpose sentence embedding learning methods, which are not combining multiple source embeddings as required by the meta-embedding learning methods we consider in this paper.


\subsection{Baselines Methods}
\label{sec:baselines}

We compare our proposed method against several baseline methods, which we describe next.
\begin{description}[style=unboxed,leftmargin=0cm]

    \item [Single Source Embeddings (SSE):]
    We create sentence embeddings from each source embedding separately by applying either max or mean pooling over the word embeddings produced by that source for the words in a sentence. 
    These baselines will demonstrate the level of performance that we can expect to obtain if we had used a single source embedding.
    This can be seen as a lower baseline for meta-embedding because one would expect a meta-embedding to always outperform the individual source embeddings used in it.
    We use both mean and max pooling given respectively by \eqref{eq:mean-pool} and \eqref{eq:max-pool} to create sentence embeddings from the source embeddings in our evaluations.
    
    \item [Concatenation (CONC):]
    In prior work of meta-embedding, it has been reported that by simply concatenating the different source embeddings of a word, one can obtain a surprisingly accurate meta-embedding for that word~\cite{yin-schutze-2016-learning,coates-bollegala-2018-frustratingly}. 
    We extend this method to create sentence-level meta-embeddings by first creating word-level meta-embeddings for each word in a given sentence by the concatenation of each source embedding of that word.
    Next, we apply max or mean pooling over those concatenated word-level meta-embeddings.
    Specifically, for a given sentence $s$, its the sentence-level meta-embeddings, ${\rm conc}_{\rm max}(s)$ and ${\rm conc}_{\rm mean}(s)$ are computed using respectively max and mean pooling methods as follows:
    \begin{align}
        {\rm conc}_{\rm max}(s) = \max_{w \in s} \bigoplus_{i=1}^n \vec{f}_i(w,s) \label{eq:conc-max} \\
        {\rm conc}_{\rm mean}(s) = \frac{1}{|s|}\sum_{w \in s} \bigoplus_{i=1}^n \vec{f}_i(w,s) \label{eq:conc-mean}
    \end{align}
    Here, $\bigoplus$ denotes the vector concatenation.
    Note that token pooling methods are applied \emph{after} the concatenation operation.
    
    One disadvantage of concatenation as a meta-embedding method is that because it adds up the dimensionalities of the source embeddings, when the number of sources increases or when the dimensionality of the source embeddings increases, thereby creating high-dimensional meta-embeddings.
    High-dimensional meta-embeddings require large storage spaces and are also slower to conduct operations such as computing cosine similarity between words/sentences.
    This is particularly problematic when creating meta-embeddings of contextualised embeddings produced by NLMs such as BERT and RoBERTa, which tend to have larger dimensionalities (e.g. 1024, 2048 etc.).
    
    \item [Averaging (AVG):]
    Averaging is also a simple method to compute meta-embeddings that has performed surprisingly well in downstream tasks~\cite{coates-bollegala-2018-frustratingly,poerner-etal-2020-sentence}.
    Unlike concatenation, averaging \emph{does not} increases the dimensionality of the meta-embedding when the number of source embeddings increases.
    This overcomes the computational and storage related disadvantages of concatenation described above.
    Specifically, for a given sentence $s$, its the sentence-level meta-embeddings, ${\rm avg}_{\rm max}(s)$ and ${\rm avg}_{\rm mean}(s)$ are computed using respectively max and mean pooling methods as follows:
    \begin{align}
        {\rm avg}_{\rm max}(s) = \max_{w \in s} \sum_{i=1}^n \vec{f}_i(w,s) \label{eq:avg-max} \\
        {\rm avg}_{\rm mean}(s) = \frac{1}{|s|}\sum_{w \in s} \sum_{i=1}^n \vec{f}_i(w,s) \label{eq:avg-mean}
    \end{align}
    
    Although one cannot simply add (or average) vectors that lie in different vector spaces, \newcite{coates-bollegala-2018-frustratingly} showed that in the case of static word embeddings, such operations can be carried out because the source word embeddings are found to be approximately orthogonal in practice.
    However, this requirement might not be readily satisfied by a set of arbitrary source embeddings and prior work have trained orthogonal projections to explicitly enforce this orthogonality requirement, which has improved the performance of the resultant meta-embeddings~\cite{jawanpuria-etal-2020-learning,he-etal-2020-learning}.
    Moreover, all source embeddings must have equal dimensionalities in order for us to compute their average.
    When this requirement is not satisfied, prior work have padded zeros as necessary prior to computing the average.
    Note that our proposal to learn projection matrices for individual source embeddings as given in \eqref{eq:w-attention} automatically solves both of those issues.
    Specifically, the projection matrices $\mat{A}_i$ convert source embeddings with potentially different dimensionalities to the same meta-embedding space (thereby obviating the need for zero padding), and simultaneously learn orthogonal projections.

    \item [SVD:] Singular Value Decomposition was used as a method for creating sentence-level meta-embeddings by \newcite{yin-schutze-2016-learning}.
    Given a matrix $\mat{A} \in \R^{d_m \times d_n}$, its SVD is given by $\mat{A} = \mat{V}\mat{S}\mat{U}\T$ where $\mat{V}, \mat{S}, \mat{U} \in \R^{d_m \times d_m}, \R^{d_m \times d_n}, \R^{d_n \times d_n}$.
    We follow \newcite{poerner-etal-2020-sentence}  and apply SVD on the matrix $\mat{A}$ formed by row-wise zero-mean concatenation matrix of the source embeddings. 
    We then use $\mat{U}\T$  as the meta-sentence embedding. 
    SVD overcomes the increasing dimensionality problem associated with CONC because the dimensionality is reduced in the SVD process by limiting to the largest eigenvectors.
    
    \item [GCCA:] Generalised Canonical Correlation Analysis~\cite{GCCA} (GCCA) was used as a method for creating sentence-level meta-embeddings by \newcite{poerner-etal-2020-sentence}.
    Given two random vectors $\vec{x}$ and $\vec{y}$, the Canonical Correlation Analysis (CCA) finds linear projections $\vec{\lambda}$ and $\vec{\mu}$ such that $\vec{\lambda}\T\vec{x}$ and $\vec{\mu}\T\vec{y}$ are maximally correlated.
    GCCA extends this approach to three or more random vectors. 
    We follow \newcite{poerner-etal-2020-sentence} and use the generalised eigenvalue decomposition-based approach proposed by \newcite{Bach:2002} to compute GCCA.
    GCCA is the current state-of-the-art method for creating meta sentence embeddings.
    
    \item [Supervised Meta-Embedding (SUP):] \label{sec:sup} 
    Note that all of the above-mentioned baselines are all unsupervised meta-embedding methods, similar to our proposed method.
    To simulate the level of performance that we can hope to achieve if we had access to some labelled data for the target task, we propose a supervised baseline as follows.
    For this purpose we consider predicting semantic textual similarity between two given sentence as the target task.
    First, given a pair of sentences $(s, s')$ and their human similarity rating $r(s,s')$, we compute a meta sentence embedding separately for $s$ and $s'$ using either mean pooling (given by \eqref{eq:mean-pool}) and max pooling (given by \eqref{eq:max-pool}).
    Let us denote the meta sentence embedding for $s$ and $s'$ respectively by $\vec{m}(s)$ and $\vec{m}(s')$.
    We then compute the cosine similarity, $\cos(\vec{m}(s), \vec{m}(s'))$ using $\vec{m}(s)$ and $\vec{m}(s')$ and minimise the squared loss between the predicted cosine similarity and the human similarity rating, $r(s,s')$, between $s$ and $s'$ as given by \eqref{eq:sup-loss}.
    \begin{align}
        \label{eq:sup-loss}
        &L(\{\mat{A}_i\}_{i=1}^{n}, \{\alpha_i\}_{i=1}^{n}) \nonumber \\ 
        =& \left(\cos(\vec{m}(s), \vec{m}(s')) - r(s,s')\right)^2
        \end{align}
    Note that projection matrices $\{\mat{A}_i\}_{i=1}^{n}$ and weights $\{\alpha_i\}_{i=1}^{n}$ enter into \eqref{eq:sup-loss} via \eqref{eq:w-attention} and get updated during backpropagation.
    
    We use the training sentence pairs in the STS-B dataset~\cite{cer-etal-2017-semeval} and minimise the loss given by \eqref{eq:sup-loss} over all such sentence pairs.
    The original human similarity ratings in STS-B are in the range $[0,5]$.
    We linearly transform both cosine similarities and human similarity ratings to $[-1,1]$ by $\frac{x - \min(x)}{\max(x)-\min(x)}$ before computing the loss given in \eqref{eq:sup-loss}.
\end{description}

\subsection{Training Details}
\label{sec:hyperparameters}

To train UNSUP, we use the set of sentences that appear in the sentence pairs in the STS-B training split.
Note that we \emph{do not} use any human similarity ratings for the sentence pairs in the STS-B training split during the training of UNSUP.
In fact, we do not even require pairs of sentences as they appear in STS-B for training UNSUP.
We randomly pick two sentences that appear in (could be the same or different sentence pairs) and compute the loss given by \eqref{eq:unsup-loss},
and update the projection matrices and weights such that the loss is minimised.

To train SUP, we use the sentence pairs and their human similarity ratings from the training sentence pairs in STS-B dataset.

We use SGD optimiser with weight decay $10^{-4}$, learning rate $10^{-2}$ for both UNSUP and SUP methods, and a batch size of 512.
We apply an early stopping as a regularisation technique such that, if the Pearson correlation does not improve for more than five consecutive epochs, computed over the development set in STS-B we stop further training.
All the parameters SUP and UNSUP are randomly initialised in the $[-1,1]$ range.

We use Quadro RTX 8000 with 48GB GPU-RAM for training. On this machine, it takes eight hours to complete the training of the proposed sentence-level meta-embedding method.

\begin{table}[t]
\begin{threeparttable}
\caption{Results of meta-sentence embeddings by using BERT and RoBERTa from sBERT \protect\cite{reimers-2019-sentence-bert}.
We show both Pearson $r$ and Spearman $\rho$ correlation coefficients, separated by /.}
\label{table:main_results}
\small
\begin{tabular}{lrrr}\toprule
Method & STS-15 & STS-16 & STS-B  \\ \midrule
SSE & & & \\
-BERT\tnote{a} & 87.15/87.34 & 81.98/83.05 & 82.40/82.74 \\ 
-BERT\tnote{b} & 88.29/88.22 & 83.24/84.02 & 83.66/84.05 \\ 
-RoBERTa\tnote{a} & 75.94/78.69 & 72.96/75.20 & 82.46/83.75 \\ 
-RoBERTa\tnote{b} & 87.72/87.88 & 84.55/84.93 & 82.46/83.26 \\ 
\midrule
Meta & & & \\
CONC\tnote{a} & 82.46/83.86 & 78.33/79.88 & 75.12/76.21  \\ 
CONC\tnote{b} & 88.64/88.59 & 84.27/84.88 & 84.69/85.14  \\ 
AVG\tnote{a} & 82.23/83.72 & 77.99/79.68 & 83.89/84.53  \\ 
AVG\tnote{b} & 88.47/88.45 & 83.97/84.63 & 84.21/84.47  \\ 
SVD & 88.66/88.65 & 84.07/84.63 & 83.98/84.61 \\ 
GCCA & 88.58/88.58 & 84.00/84.53 & 83.36/84.14 \\ 
\midrule
SUP & 89.34/89.30 & 85.11/85.72 & 65.21/64.79 \\ 
UNSUP & \textbf{88.76}/\textbf{88.85} & \textbf{85.06}/\textbf{85.33} & \textbf{85.33}/\textbf{86.08} \\ 
\bottomrule
\end{tabular}
\begin{tablenotes}
{\small \item[a] max-pooling for word-to-sentence pooling.
\item[b] mean-pooling for word-to-sentence pooling.}
\end{tablenotes}
\end{threeparttable}
\end{table}



\subsection{Results}
\label{sec:results}

\autoref{table:main_results} summarises the main results for the proposed unsupervised method (\textbf{UNSUP}) and the baseline methods described in \autoref{sec:baselines}.
We use 1024-dimensional BERT and RoBERTa from sBERT \cite{reimers-2019-sentence-bert} as the source embeddings in all the experiments.
The dimensionality in all meta-embedding methods compared in \autoref{table:main_results} are 1024, except for CONC, which is 2048 dimensional.
Where applicable, we show the results obtained for the sentence embeddings created by max and mean pooling methods.

From \autoref{table:main_results}, we see that for single source embeddings (SSE) the performance is consistently better when we use mean-pooling instead of max-pooling, except for RoBERTa in STS-B, where there is no significant difference between the two pooling methods.
In particular, we see that mean pooling to produce better sentence embeddings than max pooling for both BERT and RoBERTa source embeddings.
Given that both BERT and RoBERTa source embeddings are pretrained on sentence similarity tasks, one reason for this behaviour could be that mean pooling averages all token-level embeddings in the final layer of the MLM, thus more robust to noise compared to max pooling, which represents the meaning of a sentence using the embedding of a single token.

Next, in \autoref{table:main_results}, we compare the performance of different baseline methods described in \autoref{sec:baselines}.
We evaluate CONC and AVG using both max and mean pooling methods for creating sentence embeddings in \autoref{table:main_results}.
We see that CONC consistently reporting better performances than the best between the two individual source embedding in all datasets.
This shows that CONC is still a strong baseline not only when computing meta embeddings from static word embeddings as reported in prior work~\cite{coates-bollegala-2018-frustratingly}, but also in the case of contextualised word embeddings.

AVG on the other hand closely follows the performance of CONC but never outperforms CONC in any of the datasets.
Here again we see that the mean pooling variants of CONC and AVG outperform their max pooling counterparts.
Although SVD slightly outperforms CONC in terms of Spearman correlation in STS-15, it reports consistently lower performance than CONC in all other settings.
This shows that although SVD can reduce the dimensionality compared to CONC, the performance drops due to possible information loss during the dimensionality reduction process.
GCCA consistently under performs compared to SVD in all datasets.



The proposed \textbf{UNSUP} and the supervised baseline (\textbf{SUP}) are shown at the bottom two rows in \autoref{table:main_results}.
We see that SUP outperforms all baselines including CONC in both STS-15 and STS-16 datasets but not on STS-B.
Therefore, the supervised meta-embedding learning method described in \autoref{sec:baselines} can be considered as a strong baseline for incorporating human similarity ratings into the meta-embedding learning process.
However, UNSUP outperforms all methods compared in \autoref{table:main_results} and obtains the best performance on all datasets in terms of both Spearman and Pearson correlation coefficients.
In particular, UNSUP outperforms the current state-of-the-art meta sentence embedding method, GCCA, proposed by \newcite{poerner-etal-2020-sentence}. 
The 95\% confidence intervals for the Pearson's $r$ for \textbf{UNSUP} on STS15, STS16 and STS-B are respectively [88.29, 89.23], [84.45, 85.67], and [83.88, 86.78], which overlap with several other methods indicating that the improvements are not satistically significant.
The supervision considered by \textbf{SUP} is limited to the updating the projection matrices such that the similarity ratings in training datasets could be accurately predicted.
However, there exist alternative methods to provide supervision for sentence-level meta-embeddings such as by fine-tuning the meta-embedding models on sentence pairs in benchmark datasets.
However, we consider this to be beyond the scope of the current work, which focuses on unsupervised sentence-level meta-embedding methods that can be used in the absence of any human supervision.

\subsection{Effect of Weighting and Pooling}
\label{sec:ablation}

\begin{table}[t]
\caption{Ablation of attention and pooling for meta sentence embeddings with BERT and RoBERTa used as the source embeddings. 
The column corresponding to Pooling denotes the pooling method used for creating sentence-level meta-embeddings from word-level meta-embeddings as given by \eqref{eq:mean-pool} and \eqref{eq:max-pool}. 
Weight column denotes the weights being used. Specifically, w/ represents  using the weights $\alpha_i$ in \eqref{eq:w-attention}, whereas w/o represents not using the weights as in \eqref{eq:wo-attention}.
Correlation coefficients are evaluated on STS-B dataset.}
\label{table:experimental_settings}
\begin{tabular}{l|rrrr}\toprule
Method & Weight & Pooling & Pearson & Spearman  \\
\midrule
UNSUP & w/ & mean & 75.08 & 75.03 \\ 
UNSUP & w/ & max & \textbf{85.33} & \textbf{86.08} \\ 
UNSUP & w/o & mean & 80.06 & 80.41 \\ 
UNSUP & w/o & max & 81.98 & 83.00 \\ 
\bottomrule
\end{tabular}
\end{table}

We study the effect of using weights $\alpha_i$ with max and mean pooling methods in \autoref{table:experimental_settings}.
Specifically, w/ represents  using the weights $\alpha_i$ in \eqref{eq:w-attention}, whereas w/o represents not using the weights as in \eqref{eq:wo-attention}.
From \autoref{table:experimental_settings}, we see that the best performance is obtained with UNSUP when we use max pooling and with weighting.

\section{Conclusion}
We propose an unsupervised method to create sentence-level meta-embeddings from multiple contextualised word embeddings.
Our proposed method outperformed several competitive baselines and a previous proposed state-of-the-art sentence-level unsupervised meta-embedding method.
In particular, both weighted linear transformation and pooling operations contributed to the optimal performance of the proposed method.

\section*{Bibliographic References}
\bibliographystyle{lrec2022-bib}
\bibliography{Embed}

\begin{thebibliography}{}

\bibitem[\protect\citename{Agirre \bgroup et al.\egroup
  }2015]{agirre-etal-2015-semeval}
Agirre, E., Banea, C., Cardie, C., Cer, D., Diab, M., Gonzalez-Agirre, A., Guo,
  W., Lopez-Gazpio, I., Maritxalar, M., Mihalcea, R., Rigau, G., Uria, L., and
  Wiebe, J.
\newblock (2015).
\newblock {S}em{E}val-2015 task 2: Semantic textual similarity, {E}nglish,
  {S}panish and pilot on interpretability.
\newblock In {\em Proceedings of the 9th International Workshop on Semantic
  Evaluation ({S}em{E}val 2015)}, pages 252--263, Denver, Colorado, June.
  Association for Computational Linguistics.

\bibitem[\protect\citename{Agirre \bgroup et al.\egroup
  }2016]{agirre-etal-2016-semeval}
Agirre, E., Banea, C., Cer, D., Diab, M., Gonzalez-Agirre, A., Mihalcea, R.,
  Rigau, G., and Wiebe, J.
\newblock (2016).
\newblock {S}em{E}val-2016 task 1: Semantic textual similarity, monolingual and
  cross-lingual evaluation.
\newblock In {\em Proceedings of the 10th International Workshop on Semantic
  Evaluation ({S}em{E}val-2016)}, pages 497--511, San Diego, California, June.
  Association for Computational Linguistics.

\bibitem[\protect\citename{Bach and Jordan}2002]{Bach:2002}
Bach, F.~R. and Jordan, M.~I.
\newblock (2002).
\newblock Kernel independent component analysis.
\newblock {\em Journal of Machine Learning Research}, 3:1--48, July.

\bibitem[\protect\citename{Bahdanau \bgroup et al.\egroup }2015]{Bahdanau:2014}
Bahdanau, D., Cho, K., and Bengio, Y.
\newblock (2015).
\newblock Neural machine translation by jointly learning to align and
  translate.
\newblock {\em Proc. of ICLR}.

\bibitem[\protect\citename{Bao and Bollegala}2018]{Bao:COLING:2018}
Bao, C. and Bollegala, D.
\newblock (2018).
\newblock Learning word meta-embeddings by autoencoding.
\newblock In {\em Proc. of COLING}, pages 1650--1661.

\bibitem[\protect\citename{Bollegala \bgroup et al.\egroup
  }2018]{Bollegala:IJCAI:2018}
Bollegala, D., Hayashi, K., and Kawarabayashi, K.-i.
\newblock (2018).
\newblock Think globally, embed locally --- locally linear meta-embedding of
  words.
\newblock In {\em Proc. of IJCAI-EACI}, pages 3970--3976.

\bibitem[\protect\citename{Cer \bgroup et al.\egroup
  }2017]{cer-etal-2017-semeval}
Cer, D., Diab, M., Agirre, E., Lopez-Gazpio, I., and Specia, L.
\newblock (2017).
\newblock {S}em{E}val-2017 task 1: Semantic textual similarity multilingual and
  crosslingual focused evaluation.
\newblock In {\em Proceedings of the 11th International Workshop on Semantic
  Evaluation ({S}em{E}val-2017)}, pages 1--14, Vancouver, Canada, August.
  Association for Computational Linguistics.

\bibitem[\protect\citename{Coates and
  Bollegala}2018a]{coates-bollegala-2018-frustratingly}
Coates, J. and Bollegala, D.
\newblock (2018a).
\newblock Frustratingly easy meta-embedding {--} computing meta-embeddings by
  averaging source word embeddings.
\newblock In {\em Proceedings of the 2018 Conference of the North {A}merican
  Chapter of the Association for Computational Linguistics: Human Language
  Technologies, Volume 2 (Short Papers)}, pages 194--198, New Orleans,
  Louisiana, June. Association for Computational Linguistics.

\bibitem[\protect\citename{Coates and Bollegala}2018b]{Coates:NAACL:2018}
Coates, J. and Bollegala, D.
\newblock (2018b).
\newblock Frustratingly easy meta-embedding -- computing meta-embeddings by
  averaging source word embeddings.
\newblock In {\em Proc. of NAACL-HLT}, pages 194--198.

\bibitem[\protect\citename{Conneau and Kiela}2018]{Conneau:2018ac}
Conneau, A. and Kiela, D.
\newblock (2018).
\newblock {S}ent{E}val: {A}n {E}valuation {T}oolkit for {U}niversal {S}entence
  {R}epresentations.

\bibitem[\protect\citename{Devlin \bgroup et al.\egroup }2019]{BERT}
Devlin, J., Chang, M.-W., Lee, K., and Toutanova, K.
\newblock (2019).
\newblock {BERT}: Pre-training of deep bidirectional transformers for language
  understanding.
\newblock In {\em Proc. of NAACL-HLT}, pages 4171--4186.

\bibitem[\protect\citename{Goikoetxea \bgroup et al.\egroup
  }2016]{AAAI:2016:Goikoetxea}
Goikoetxea, J., Agirre, E., and Soroa, A.
\newblock (2016).
\newblock Single or multiple? combining word representations independently
  learned from text and wordnet.
\newblock In {\em Proc. of AAAI}, pages 2608--2614.

\bibitem[\protect\citename{He \bgroup et al.\egroup
  }2020a]{he-etal-2020-learning}
He, J., Tsiolis, K., Kenyon-Dean, K., and Cheung, J. C.~K.
\newblock (2020a).
\newblock Learning efficient task-specific meta-embeddings with word prisms.
\newblock In {\em Proceedings of the 28th International Conference on
  Computational Linguistics}, pages 1229--1241, Barcelona, Spain (Online),
  December. International Committee on Computational Linguistics.

\bibitem[\protect\citename{He \bgroup et al.\egroup }2020b]{He:2020}
He, J., Tsiolis, K., Kenyon-Dean, K., and Cheung, J. C.~K.
\newblock (2020b).
\newblock {L}earning {E}fficient {T}ask-{S}pecific {M}eta-{E}mbeddings with
  {W}ord {P}risms.
\newblock In {\em Proc. of COLING}.

\bibitem[\protect\citename{Horst}1961]{GCCA}
Horst, P.
\newblock (1961).
\newblock Generalized canonical correlations and their applications to
  experimental data.
\newblock In {\em Clinical Psychology}, volume~17, pages 331--347.

\bibitem[\protect\citename{Jawanpuria \bgroup et al.\egroup
  }2020]{jawanpuria-etal-2020-learning}
Jawanpuria, P., N~T~V, S.~D., Kunchukuttan, A., and Mishra, B.
\newblock (2020).
\newblock Learning geometric word meta-embeddings.
\newblock In {\em Proc. of RepL4NLP}, pages 39--44.

\bibitem[\protect\citename{Kiela \bgroup et al.\egroup
  }2018]{kiela-etal-2018-dynamic}
Kiela, D., Wang, C., and Cho, K.
\newblock (2018).
\newblock Dynamic meta-embeddings for improved sentence representations.
\newblock In {\em Proc. of EMNLP}, pages 1466--1477.

\bibitem[\protect\citename{Liu \bgroup et al.\egroup }2019]{RoBERTa}
Liu, Y., Ott, M., Goyal, N., Du, J., Joshi, M., Chen, D., Levy, O., Lewis, M.,
  Zettlemoyer, L., and Stoyanov, V.
\newblock (2019).
\newblock {R}o{B}{E}{R}{T}a: {A} {R}obustly {O}ptimized {B}{E}{R}{T}
  {P}retraining {A}pproach.

\bibitem[\protect\citename{Mikolov \bgroup et al.\egroup
  }2013a]{DBLP:journals/corr/abs-1301-3781}
Mikolov, T., Chen, K., Corrado, G., and Dean, J.
\newblock (2013a).
\newblock Efficient estimation of word representations in vector space.
\newblock In Yoshua Bengio et~al., editors, {\em 1st International Conference
  on Learning Representations, {ICLR} 2013, Scottsdale, Arizona, USA, May 2-4,
  2013, Workshop Track Proceedings}.

\bibitem[\protect\citename{Mikolov \bgroup et al.\egroup
  }2013b]{NIPS2013_9aa42b31}
Mikolov, T., Sutskever, I., Chen, K., Corrado, G.~S., and Dean, J.
\newblock (2013b).
\newblock Distributed representations of words and phrases and their
  compositionality.
\newblock In C.~J.~C. Burges, et~al., editors, {\em Advances in Neural
  Information Processing Systems}, volume~26. Curran Associates, Inc.

\bibitem[\protect\citename{O'Neill and Bollegala}2018]{Neill:2018}
O'Neill, J. and Bollegala, D.
\newblock (2018).
\newblock {A}ngular-{B}ased {W}ord {M}eta-{E}mbedding {L}earning.

\bibitem[\protect\citename{Pennington \bgroup et al.\egroup
  }2014]{Pennington:EMNLP:2014}
Pennington, J., Socher, R., and Manning, C.~D.
\newblock (2014).
\newblock Glove: global vectors for word representation.
\newblock In {\em Proc. of EMNLP}, pages 1532--1543.

\bibitem[\protect\citename{Poerner \bgroup et al.\egroup
  }2020]{poerner-etal-2020-sentence}
Poerner, N., Waltinger, U., and Sch{\"u}tze, H.
\newblock (2020).
\newblock Sentence meta-embeddings for unsupervised semantic textual
  similarity.
\newblock In {\em Proc. of ACL}, pages 7027--7034.

\bibitem[\protect\citename{Reimers and
  Gurevych}2019]{reimers-2019-sentence-bert}
Reimers, N. and Gurevych, I.
\newblock (2019).
\newblock Sentence-bert: Sentence embeddings using siamese bert-networks.
\newblock In {\em Proceedings of the 2019 Conference on Empirical Methods in
  Natural Language Processing}. Association for Computational Linguistics, 11.

\bibitem[\protect\citename{Sennrich \bgroup et al.\egroup }2016]{Sennrich:2016}
Sennrich, R., Haddow, B., and Birch, A.
\newblock (2016).
\newblock Neural machine translation of rare words with subword units.
\newblock In {\em Proceedings of the 54th Annual Meeting of the Association for
  Computational Linguistics (Volume 1: Long Papers)}, pages 1715--1725, Berlin,
  Germany, August. Association for Computational Linguistics.

\bibitem[\protect\citename{Wolf \bgroup et al.\egroup
  }2020]{wolf-etal-2020-transformers}
Wolf, T., Debut, L., Sanh, V., Chaumond, J., Delangue, C., Moi, A., Cistac, P.,
  Rault, T., Louf, R., Funtowicz, M., Davison, J., Shleifer, S., von Platen,
  P., Ma, C., Jernite, Y., Plu, J., Xu, C., Scao, T.~L., Gugger, S., Drame, M.,
  Lhoest, Q., and Rush, A.~M.
\newblock (2020).
\newblock Transformers: State-of-the-art natural language processing.
\newblock In {\em Proceedings of the 2020 Conference on Empirical Methods in
  Natural Language Processing: System Demonstrations}, pages 38--45, Online,
  October. Association for Computational Linguistics.

\bibitem[\protect\citename{Wu \bgroup et al.\egroup }2020]{wu2020task}
Wu, X., Cai, Y., Kai, Y., Wang, T., and Li, Q.
\newblock (2020).
\newblock Task-oriented domain-specific meta-embedding for text classification.
\newblock In {\em Proc. of EMNLP}, pages 3508--3513.

\bibitem[\protect\citename{Xie \bgroup et al.\egroup }2019]{xie2019dynamic}
Xie, Y., Hu, Y., Xing, L., and Wei, X.
\newblock (2019).
\newblock Dynamic task-specific factors for meta-embedding.
\newblock In {\em International Conference on Knowledge Science, Engineering
  and Management}, pages 63--74. Springer.

\bibitem[\protect\citename{Yin and Sch{\"u}tze}2016a]{Yin:ACL:2016}
Yin, W. and Sch{\"u}tze, H.
\newblock (2016a).
\newblock Learning word meta-embeddings.
\newblock In {\em Proc. of ACL}, pages 1351--1360.

\bibitem[\protect\citename{Yin and
  Sch{\"u}tze}2016b]{yin-schutze-2016-learning}
Yin, W. and Sch{\"u}tze, H.
\newblock (2016b).
\newblock Learning word meta-embeddings.
\newblock In {\em Proceedings of the 54th Annual Meeting of the Association for
  Computational Linguistics (Volume 1: Long Papers)}, pages 1351--1360, Berlin,
  Germany, August. Association for Computational Linguistics.

\end{thebibliography}

\end{document}